\def\year{2019}\relax
\documentclass[letterpaper]{article} %DO NOT CHANGE THIS
\usepackage{aaai19}  %Required
\usepackage{times}  %Required
\usepackage{helvet}  %Required
\usepackage{courier}  %Required
\usepackage{url}  %Required
\usepackage{graphicx}  %Required
\usepackage{multirow}
\usepackage{algorithm}
\usepackage{algorithmic}
\usepackage{amsfonts,amssymb}
\usepackage{booktabs}
\begin{document}
% The file aaai.sty is the style file for AAAI Press
% proceedings, working notes, and technical reports.
%
\title{Building Sequential Inference Models for End-to-End Response Selection}
\author{Jia-Chen Gu$^1$, Zhen-Hua Ling$^1$, Yu-Ping Ruan$^1$ \and Quan Liu$^{1,2}$ \\
$^1$National Engineering Laboratory for Speech and Language Information Processing, \\
    University of Science and Technology of China, Hefei, China \\
$^2$iFLYTEK Research, Hefei, China \\
gujc@mail.ustc.edu.cn, zhling@ustc.edu.cn, ypruan@mail.ustc.edu.cn, quanliu@ustc.edu.cn \\
}

%\author{Jia-Chen Gu \\
%  University of Science and \\
%   Technology of China \\
%  {\tt gujc@mail.ustc.edu.cn} \\ \And
%  Zhen-Hua Ling \\
%  University of Science and \\
%   Technology of China \\
%  {\tt zhling@ustc.edu.cn} \\ \And
%  Yu-Ping Ruan \\
%  University of Science and \\
%   Technology of China \\
%  {\tt ypruan@mail.ustc.edu.cn} \\ \And
%  Quan Liu \\
%  iFLYTEK \\
%  Research \\
%  {\tt quanliu@iflytek.com} \\
%}

%\nocopyright
\maketitle

\begin{abstract}
 This paper presents an end-to-end response selection model for Track 1 of the 7th Dialogue System Technology Challenges (DSTC7). This task focuses on selecting the correct next utterance from a set of candidates given a partial conversation. We propose an end-to-end neural network based on enhanced sequential inference model (ESIM) for this task. Our proposed model differs from the original ESIM model in the following four aspects. First, a new word representation method which combines the general pre-trained word embeddings with those estimated on the task-specific training set is adopted in order to address the challenge of out-of-vocabulary (OOV) words. Second, an attentive hierarchical recurrent encoder (AHRE) is designed which is capable to encode sentences hierarchically and generate more descriptive representations by aggregation. Third, a new pooling method which combines multi-dimensional pooling and last-state pooling is used instead of the simple combination of max pooling and average pooling in the original ESIM. Last, a modification layer is added before the softmax layer to emphasize the importance of
 %modify the matching degree between the context and the response with only
 the last utterance in the context for response selection. In the released evaluation results of DSTC7, our proposed method ranked second on the Ubuntu dataset and third on the Advising dataset in subtask 1 of Track 1.
\end{abstract}

\section{Introduction}
 Building dialogue systems that can converse naturally with humans is a challenging yet intriguing problem of artificial intelligence. Recently, human-computer conversation has attracted increasing attention due to its promising potentials and alluring commercial values. According to the applications, dialogue systems can be roughly divided into two categories : (1) task-oriented systems and (2) non-task-oriented systems (also known as chatbots). Task-oriented systems aim to assist the user to complete certain tasks (e.g. booking accommodations and restaurants). Non-task-oriented systems aim to engage users in human-computer conversations in the open domain and attract lots of research attentions because they target on unstructured dialogues without a priori logical representation for the information exchanging during the conversation.

 Existing approaches to dialogue response generation includes generation-based methods \cite{shang2015neural,serban2016building} and retrieval-based methods \cite{zhou2016multi,wu2017sequential}. Generation-based models maximize the probability of generating a response given the previous dialogue. This approach enables the incorporation of rich context when mapping between consecutive dialogue turns. Retrieval-based methods select a proper response for the current conversation from a repository with response selection algorithms, and have the advantage of producing informative and fluent responses. Track 1 of the 7th Dialogue System Technology Challenges (DSTC7) is a kind of retrieval-based task which selects the correct response from a large set of candidates. The set used in this track contains more candidates than many other datasets. Some candidates are also similar which increases the difficulty of making right decisions.

 %To solve response selection task well, one key to it is
 The techniques of word embeddings and sentence embeddings  are important to response selection as well as many other natural language processing (NLP) tasks.
 The context and the response must be projected to a vector space appropriately in order to capture the relationships between them,
 %which will spontaneously help many downstream parts.
 which are essential for following procedures.
 Recently there has been a growing interest in models for word-level \cite{mikolov2013distributed,pennington2014glove,dong2018enhance} and sentence-level \cite{wang2017bilateral,chen2017enhanced} representations using neural networks,
 %Distributed representations of words and sentences in a vector space
 which helped classification or inference algorithms to achieve better performance in many NLP tasks. %by grouping similar words, and what's more,
 %and they are also the basic of sentence embeddings.
 %Sentence embeddings are distributed representations of natural language sentences which encode the meaning of the sentences in a neural network representation.
 %Projecting a sentence to a vector space can help us perform more downstream NLP tasks.

 Another key technique to the response selection task lies in context-response matching. %Matching and inference are core functions of both human and artificial intelligence.
 Modeling the semantic matching degree between two sentences is challenging.
 The enhanced sequential inference model (ESIM) \cite{chen2017enhanced} was proposed to measure the relationship between a pair of sentences in natural language inference (NLI) tasks. This model described the interactions between two sentences by sequential encoding and attention-based alignment.
 %shows great performance on matching pair of sentences because it conducts interactions between pair of sentences which can help capture the information from the other. The approach only relies on alignment and is fully computationally decomposable with respect to the input text. For the effectiveness of ESIM, we set it as our baseline.
 Considering the good performance and decomposable implementation of ESIM, it is adopted as our baseline model for response selection.

 This paper introduces the end-to-end response selection method developed by us for subtask 1 of Track 1 in DSTC7.
 We propose to improve the original ESIM model for response selection from the following four aspects.
 \begin{itemize}
%  \item We used a new word representation method to address the challenge of large number of out-of-vocabulary words (OOV) words.
%  \item We proposed an attentive hierarchical recurrent encoder (AHRE) to encode the sentence at multi-level hierarchically to generate more descriptive representations.
%  \item The pooling method in the original ESIM is replaced with a combination of multi-dimensional pooling and last-state pooling.
%  \item A modification layer is employed to modify the matching degree between the context and the response with only the last utterance in a context and the response.
  \item A new word representation method which combines the general pre-trained word embeddings with those estimated on the task-specific training set is adopted in order to address the challenge of out-of-vocabulary (OOV) words.
  \item An attentive hierarchical recurrent encoder (AHRE) is designed which encodes sentences hierarchically and generates sentence representations by aggregation.
  \item A new pooling method which combines multi-dimensional pooling and last-state pooling is used instead of the simple combination of max pooling and average pooling in the original ESIM.
  \item A modification layer is added before the softmax layer to emphasize the importance of the last utterance in the context for response selection.
 \end{itemize}
 As shown in the released challenge results, our proposed model ranked second on the Ubuntu dataset and ranked third on the Advising dataset in subtask 1 of Track 1.
 In the following sections, we first introduce the task descriptions of Track 1 in DSTC7, and present the details of our proposed model.
 Then the model configurations, training settings and evaluation results are shown.
 Furthermore, the experimental results are analyzed by ablation tests. Finally we draw conclusions and give an overview of our future work.

\section{Task Description}

\begin{table}
  \centering
  \begin{tabular}{|c|l|}
  \hline
  Speaker & \multicolumn{1}{|c|}{Utterances in a dialogue} \\
  \hline
  A & Hmm, perhaps I should return to windows... \\
  \hline
  B & what kind of cd-rom do you have??  \\
  \hline
  A & It's a external USB drive....  Any ideas? \\
  \hline
  B & what cd-rom do you have?? \\
  \hline
  A & An external LG USB \\
  \hline
  \multicolumn{2}{|l|}{Candidates:} \\
  \multicolumn{2}{|l|}{1. perhaps it's not running?} \\
  \multicolumn{2}{|l|}{2. is jack running properly?} \\
  \multicolumn{2}{|l|}{...} \\
  \multicolumn{2}{|l|}{x. sry didnt see your answer, write my name so i see it.} \\
  \multicolumn{2}{|l|}{...} \\
  \multicolumn{2}{|l|}{100. odd sized card?.  why did you run mklabel?} \\
  \hline
  \multicolumn{2}{|l|}{Answer:} \\
  \multicolumn{2}{|l|}{sry didnt see your answer, write my name so i see it.} \\
  \hline
  \end{tabular}
  \caption{Dialogue example of subtask 1.}
  \label{tab1}
\end{table}

The DSTC7 Track 1 organizers provided two datasets \cite{kummerfeld2018analyzing}. One is Ubuntu Dialogue Corpus which contains dialogues between Ubuntu users for the purpose of solving an Ubuntu user¡¯s posted problem and the other is Advising Data which consists of dialogues between a student and a advisor for the purpose of guiding the student to pick courses.

The task is divided into 5 subtasks and a participant may participate in one, several, or all the subtasks. Participants are required to meet different goals for different subtasks such as selecting the next utterance from the given 100 candidates or 120k candidates, selecting the next utterance with the set of paraphrases, selecting the next utterance with a candidate pool which might not include the correct next utterance, and selecting the next utterance with a model incorporating external knowledge.
Each subtask has its corresponding dataset and each dialogue in it has its corresponding response candidates together with the correct answer.
%From our perspective subtask 1 is baseline and core of all subtasks.
Due to limited time and manpower, we only participate subtask 1 of this track, which aims to select the next utterance from a candidate set of 100 utterances.
An example dialogue and its candidates is shown in Table~\ref{tab1}.

\section{Model Description}
\begin{figure}%[H]
\centering
\includegraphics[width=6.5cm]{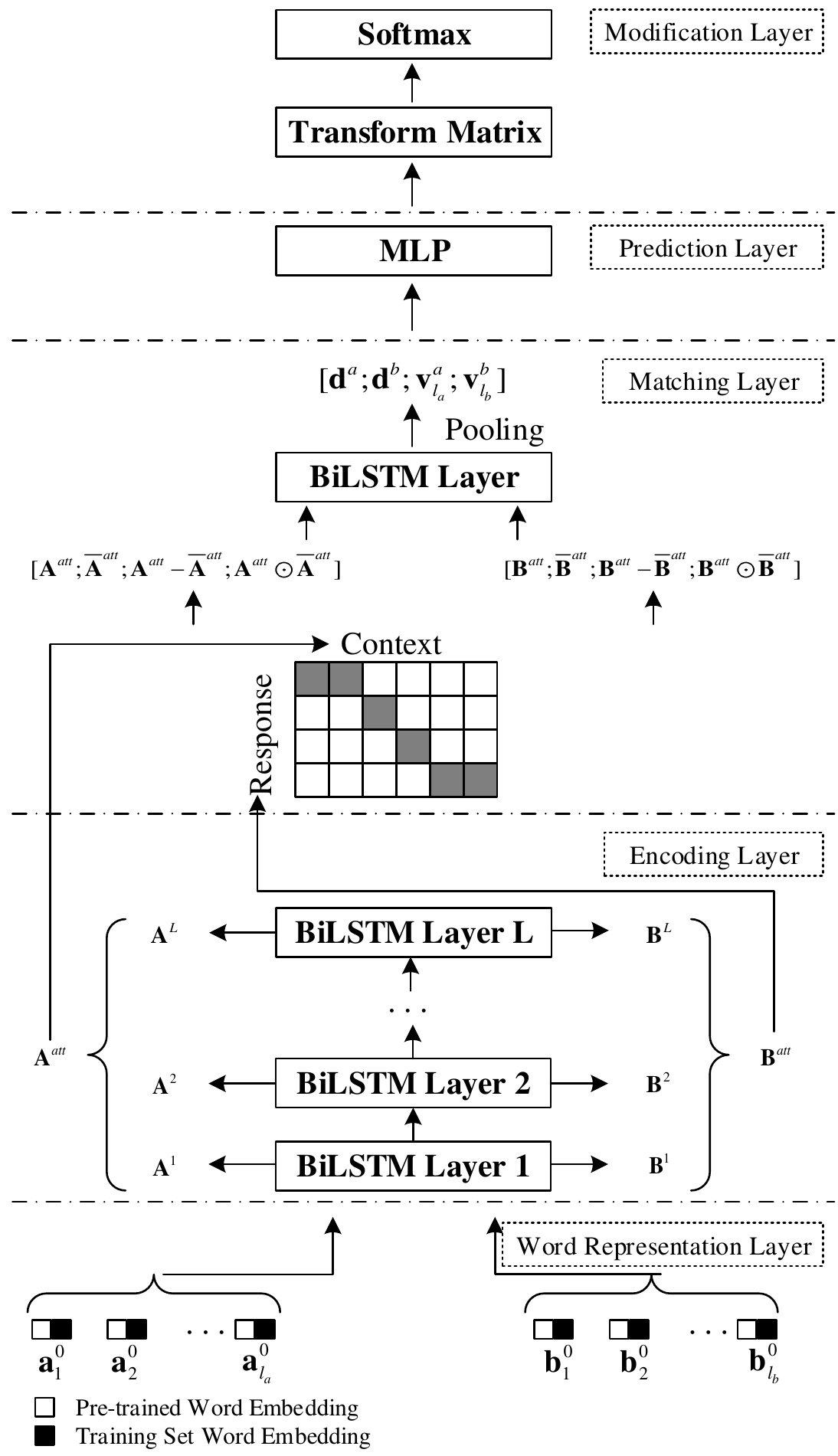}
\caption{Diagram of our proposed model.}
\label{fig1}
\end{figure}

Our proposed model is composed of five components: Word Representation Layer, Encoding Layer, Matching Layer, Prediction Layer and Modification Layer. Figure~\ref{fig1} shows the diagram of the model architecture. Details about each layer are described in this section.

  \subsection{Word Representation Layer}
  One challenge of modeling dialogue is the large number of out-of-vocabulary words. To address this issue, we adopt an algorithm \cite{dong2018enhance} which combines the general pre-trained word embedding vectors with those generated on the task-specific training set to enhance word representations.

  \subsection{Encoding Layer}
  Recurrent neural networks (RNN) \cite{mikolov2010recurrent} have been proven to be good at modeling chronological relationship in language sequences and multi-layer RNNs have achieved good performance in many NLP tasks such as neural machine translation (NMT) \cite{bahdanau2014neural} and natural language inference (NLI) \cite{chen2017enhanced}. %It is apparent that
  Encoding the sequences with deep neural networks can help capture deeper and more useful information. Typically, the outputs of the top RNN layer are regarded as the final sentence representations and the other layers are neglected. However, the lower layers can also provide useful sentence descriptions, such as
  %are usually neglected which have great abilities to model aspects of syntax\cite{hashimoto2017joint} (e.g., they can be used to do part-of-speech tagging).
  part-of-speech tagging and syntax-related ones \cite{hashimoto2017joint}.

  To make full use of the representations at all hidden layers, we propose a new sentence encoder called attentive hierarchical recurrent encoder (AHRE).
  This encoder is motivated by the method of embeddings from language models (ELMo) \cite{peters2018deep} which combines the internal states of multi-layer RNNs.
  More specifically, an AHRE learns a linear combination of the vectors stacked above each input word, which improves the performance of just using the top RNN layer in our experiments.

  Let $\textbf{A}^0=[\textbf{a}_1^0, ..., \textbf{a}_{l_a}^0]$ and $\textbf{B}^0=[\textbf{b}_1^0, ..., \textbf{b}_{l_b}^0]$ denote sequences of word representations of context and response respectively. $l_a$ and $l_b$ are token numbers in these two sequences. Both $\textbf{a}_i^0\in \mathbb{R}^l$ and $\textbf{b}_j^0 \in \mathbb{R}^l$ are \emph{l}-dimentional embedding vectors given by the word representation layer mentioned above. Furthermore, bidirectional LSTMs (BiLSTM) \cite{hochreiter1997long} are employed as our basic building blocks. In an \emph{L}-layer RNN, the $l^{th}$ layer takes the output of the ${l-1}^{th}$ layer as its input. We denote the calculations as the follows,
  \begin{equation}
  \textbf{a}_i^{l} = \textbf{BiLSTM}(\textbf{a}^{l-1}, i), i \in \{1, ..., l_a\}, l \in \{1,.., L\},
  \end{equation}
  \begin{equation}
  \textbf{b}_j^{l} = \textbf{BiLSTM}(\textbf{b}^{l-1}, j), j \in \{1, ..., l_b\}, l \in \{1,.., L\}.
  \end{equation}
  The weights for these two BiLSTMs are shared in our implementation. Due to limit space, we skip the descriptions on the basic chain LSTMs and readers can refer to \cite{hochreiter1997long} for details.

  Finally we get a set of \emph{L} representations \{$\textbf{a}^{1}, ..., \textbf{a}^{L}$\} and \{$\textbf{b}^{1}, ..., \textbf{b}^{L}$\} through the \emph{L}-layer RNNs. Typically $\textbf{a}^{L}$ or  $\textbf{b}^{L}$, i.e. the outputs of the top layer, are used as the final encoded vectors.
  Here, we propose to combine the set of representations to get enhanced representations  $\textbf{a}_i^{att}$ and $\textbf{b}_j^{att}$ by learning attention weights of all layers.
  Mathematically, we have
  \begin{equation}
  \textbf{a}_i^{att} = \sum_{l=1}^L w_l\textbf{a}_i^{l}, i \in \{1, ..., l_a\},
  \end{equation}
  \begin{equation}
  \textbf{b}_j^{att} = \sum_{l=1}^L w_l\textbf{b}_j^{l}, j \in \{1, ..., l_b\},
  \end{equation}
  where $w_l$ are softmax-normalized weights shared between context and response which need to be estimated during the training process. Our representations differ from those of traditional encoder in that ours not only considers the top layer representations but also takes the lower layer representations which may be informative into account. As a result, the representations given by our encoder are expected to capture and fuse multi-level characteristics of sentences. %, such as morphology and semantic-related ones and fuse them together.

  \subsection{Matching Layer}
  Interactions between context and response is important to provide information for deciding the matching degree between them.
  Our model follows the matching part of ESIM \cite{chen2017enhanced} which  collects local information between two sentences by attention-based alignment and is fully computationally decomposable.

  First, a soft alignment is conducted by computing the attention weight between each representation tuple $\{\textbf{a}_i^{att}, \textbf{b}_j^{att}\}$ as%between a context and a response with Equation~\ref{equ1}.
  \begin{equation}
  e_{ij} = (\textbf{a}_i^{att})^T \cdot \textbf{b}_j^{att}.
  \label{equ1}
  \end{equation}
  Then, local inference is determined by the attention weights $e_{ij}$ computed above to obtain the local relevance between a context and a response. For a word in the context, its relevant representation carried by the response is identified and composed using $e_{ij}$ as %, more specifically with Equation~\ref{equ2}.
  \begin{equation}
  \bar{\textbf{a}}_i^{att} = \sum_{j=1}^{l_b} \frac{exp(e_{ij})} {\sum_{k=1}^{l_b} exp(e_{ik})} \textbf{b}_j^{att}, i \in \{1, ..., l_a\},
  \label{equ2}
  \end{equation}
  where $\bar{\textbf{a}}_i^{att}$ is a weighted summation of $\{\textbf{b}_j^{att}\}_{j=1}^{l_b}$. Intuitively, the contents in $\{\textbf{b}_j^{att}\}_{j=1}^{l_b}$ that are relevant to $\textbf{a}_i^{att}$ are selected to form $\bar{\textbf{a}}_i^{att}$. The same calculation is performed for each word in the response as
  \begin{equation}
  \bar{\textbf{b}}_j^{att} = \sum_{i=1}^{l_a} \frac{exp(e_{ij})} {\sum_{k=1}^{l_a} exp(e_{kj})} \textbf{a}_i^{att}, j \in \{1, ..., l_b\}.
  \label{equ3}
  \end{equation}

  To further enhance the collected information, we compute the differences and the element-wise products between  $\{\textbf{A}^{att}, \bar{\textbf{A}}^{att}\}$ and between $\{\textbf{B}^{att}, \bar{\textbf{B}}^{att}\}$. %which can help sharpen local matching information.
  The difference and element-wise product are then concatenated with the original vectors %, $\textbf{A}^{att}$ and $\bar{\textbf{A}}^{att}$, or $\textbf{B}^{att}$ and $\bar{\textbf{B}}^{att}$ respectively.
  to get the enhanced representations as follows,
  \begin{equation}
  \textbf{M}_a = [\textbf{A}^{att}; \bar{\textbf{A}}^{att}; \textbf{A}^{att} - \bar{\textbf{A}}^{att} ;\textbf{A}^{att} \odot \bar{\textbf{A}}^{att}] ,
  \end{equation}
  \begin{equation}
  \textbf{M}_b = [\textbf{B}^{att}; \bar{\textbf{B}}^{att}; \textbf{B}^{att} - \bar{\textbf{B}}^{att} ;\textbf{B}^{att} \odot \bar{\textbf{B}}^{att}] .
  \end{equation}
  Then, BiLSTMs are employed to compose the enhanced local matching information $\textbf{M}_a$ and $\textbf{M}_b$ as
  \begin{equation}
  \textbf{v}_i^a = \textbf{BiLSTM}(\textbf{M}_a, i), i \in [1, ..., l_a],
  \end{equation}
  \begin{equation}
  \textbf{v}_j^b = \textbf{BiLSTM}(\textbf{M}_b, j), j \in [1, ..., l_b].
  \end{equation}
  where BiLSTMs have \emph{d} hidden units along each direction and  $\{\textbf{v}_i^a,\textbf{v}_i^b\} \in \mathbb{R}^{1\times 2d}$.

  Instead of using max pooling and average pooling in the original ESIM model, we combine multi-dimensional pooling \cite{shen2017disan} and last-state pooling to derive the final matching feature vectors from the sequences of $\textbf{v}_i^a$ and $\textbf{v}_i^b$.

  Multi-dimensional attention differs from general attention in that the logit for an input vector is not a scalar but a vector with dimensions equal to the dimensions of the input vector. This allows each dimension of the input vector to have a scalar logit, and we can perform attention in each dimension separately.
  In our model, for $\textbf{v}_i^{a}$, its logit \emph{\textbf{l}}($\textbf{v}_i^{a}$) is calculated by two linear transformations with an exponential linear units (ELU) activation function in between, i.e.,
  \begin{equation}
  \emph{\textbf{l}}(\textbf{v}_i^a) = \textbf{ELU}(\textbf{v}_i^a \textbf{W}_1^M + \textbf{b}_1^M)\textbf{W}_2^M + \textbf{b}_2^M,
  \end{equation}
  where $\{\textbf{W}_1^M, \textbf{W}_2^M\} \in \mathbb{R}^{2d\times 2d}$ and $\{\textbf{b}_1^M, \textbf{b}_2^M\} \in \mathbb{R}^{1\times 2d}$.
  Further, we have
  \begin{equation}
  \textbf{d}_i^a = softmax(\emph{\textbf{l}}(\textbf{v}_i^a)) \odot \textbf{v}_i^a ,
  \end{equation}
  \begin{equation}
  \textbf{d}^a = \sum_{i=1}^{l_a} \textbf{d}_i^a .
  \end{equation}

  The calculations of Eq. (12)-(14) are also applied to $\textbf{v}_j^{b}$ to get $\textbf{d}^b$. Finally, we combine the multi-dimentional pooling introduced above and last-state pooling to form the matching feature vector as
  \begin{equation}
  \textbf{f} = [\textbf{d}^a; \textbf{d}^b; \textbf{v}_{l_a}^{a}; \textbf{v}_{l_b}^{b}].
  \end{equation}

  \subsection{Prediction Layer}
   The matching feature vector \textbf{f} is fed into a multi-layer perception (MLP) classifier. An MLP is a feedforward neural network estimated in a supervised way using examples of features together with known labels. Here, the MLP is designed to predict whether a pair of context and response match appropriately through the matching feature \textbf{f}. Finally, the MLP returns a score $s_1$ before softmax to denote the degree of matching.

  \subsection{Modification Layer}
  At this layer, the matching score given by the prediction layer is further modified to emphasize the effect of the last utterance in the context.
  %Thus, another matching score is calculated using only the last utterances.
  %To this end, we applied a modification layer which uses the last utterance of dialogue to match the response to better reflect the matching scores.
  We denote the length of the last utterance $\textbf{u}$ as $l_u$ and its output after AHRE as $\textbf{u}^{att}$. A last-state pooling is employed over it to get its representation $\textbf{u}^{att}_{l_u}$. A transform matrix is applied to compute another matching score $s_2$ and the final score $s$ is the combination of $s_1$ and $s_2$ with a scalar weight
   \begin{equation}
   s_2 = (\textbf{u}^{att}_{l_u})^T \cdot \textbf{M} \cdot \textbf{b}^{att}_{l_b} ,
   \end{equation}
   \begin{equation}
   s = s_1 + w \cdot s_2,
   \end{equation}
   where $\textbf{M}$ and $w$ are both parameters need to be estimated during training. Finally, a softmax layer is applied to the score $s$ to predict the correct answer among all candidates. %The multi-class cross-entropy loss to select the correct response among a set candidates.
   All model parameters are estimated in an end-to-end way by minimizing the multi-class cross-entropy loss on training set.

\section{Experiments}

  \subsection{Dataset}
  There were two datasets provided by the subtask 1. Both of them provided 100k training dialogues and each was equipped with 100 candidates. They are different in the development dataset size, test dataset size and vocabulary size. Specifically, the Ubuntu dialogue has 5k development dialogues and the vocabulary size is 113k, while the Advising dialogue has only 0.5k development dialogues and the vocabulary size is only 5k.

  \subsection{Training details}
    Adam method \cite{kingma2014adam} was employed for optimization with a minibatch size of 2. The initial learning rate was 0.001 and was exponentially decayed by 0.96 every 5000 steps. The word embeddings were concatenations of 300-dimensional fixed GloVe embeddings \cite{pennington2014glove} and 100-dimensional embeddings estimated on the training set using Word2Vec \cite{mikolov2013distributed} algorithm. The word embeddings were not updated during training. All hidden states of LSTM had 200 dimensions. The number of BiLSTM layers in AHRE was 3. The MLP at the prediction layer had a hidden unit size of 256 with ReLU \cite{nair2010rectified} activation. We set the maximum context length as 160. Zeros were padded if the length was less than 160, otherwise the last 160 words were kept. We used the development dataset to select the best model for testing.

    All codes were implemented using TensorFlow framework \cite{abadi2016tensorflow} and were released to help replicate our results\footnote{https://github.com/JasonForJoy/DSTC7-ResponseSelection}.

   \subsection{Evaluation metrics}

   \begin{table*}
   \centering
   \begin{tabular}{cccccc}
   \toprule
   Development / Test         & Dataset                   & $R_{100}@1$  & $R_{100}@10$ & $R_{100}@50$  & MRR \\
   \midrule
   \multirow{4}*{Development} & Ubuntu(single)            & 0.521  & 0.817  & 0.982 & 0.616   \\

                              & Ubuntu(ensemble)          & 0.534  & 0.825  & 0.982 & 0.631   \\

                              & Advising(single)          & 0.206  & 0.556  & 0.906 & 0.323   \\

                              & Advising(ensemble)        & 0.26   & 0.626  & 0.93  & 0.377   \\
   \midrule
   \multirow{3}*{Test}        & Ubuntu(ensemble)          & 0.608  & 0.853  & 0.984 & 0.691   \\
                              & Advising-Case 1(ensemble) & 0.42   & 0.766  & 0.972 & 0.538    \\
                              & Advising-Case 2(ensemble) & 0.194  & 0.582  & 0.908 & 0.32    \\
   \bottomrule
   \end{tabular}
   \caption{Evaluation results on Ubuntu dataset and Advising dataset of subtask 1.}
   \label{tab2}
   \end{table*}

   \begin{table*}
   \centering
   \begin{tabular}{lcccc}
   \toprule
                                                       & $R_{100}@1$  & $R_{100}@10$ & $R_{100}@50$  & MRR \\
   \midrule
   Our model (single)                                  & 0.521  & 0.817  & 0.982 & 0.616  \\
   { }- Modification layer                             & 0.514  & 0.804  & 0.981 & 0.611  \\
   { }{ }- Attentive hierarchical recurrent encoder    & 0.506  & 0.799  & 0.977 & 0.602  \\
   { }{ }{ }- Multi-dimensional and last-state pooling & 0.5    & 0.791  & 0.974 & 0.598  \\
   { }{ }{ }{ }- Fixed word embedding                  & 0.488  & 0.776  & 0.969 & 0.591  \\
   \bottomrule
   \end{tabular}
   \caption{Results of ablation tests using our single model on the Ubuntu development set of subtask 1.}
   \label{tab3}
   \end{table*}

   Both datasets in the task were designed for selecting the best answer among a set of candidates for each given conversation.
   Recalls of the selected top-\emph{k} responses from 100 available candidates for each conversation (i.e., $R_{100}@k$) were employed as metrics to evaluate our model performance.
   %, which is asked to. We reported model performance as recall at k relevant texts given 100 candidates .

   We also used mean reciprocal rank (MRR) to evaluate our model performance, which is a statistic measure for evaluating any process that produces a list of possible responses to a sample of queries, ordered by probability of correctness. The reciprocal rank of a query response is the multiplicative inverse of the rank of the first correct answer, and MRR is the average of the reciprocal ranks of results for a query set \emph{Q}. It can be formulated as % Equation~\ref{equ4}:
   \begin{equation}
   MRR = \frac{1} {|Q|} \sum_{i=1}^{|Q|} \frac{1}{rank_i},
   \label{equ4}
   \end{equation}
   where $rank_i$ refers to the rank position of the first relevant document for the \emph{i}-th query.

   %These two kinds of metric are both useful for ranking task, especially $R_{100}@1$ metric is also meaningful for classifying the best relevant text. The ranking in this challenge considered
   The average of $R_{100}@10$ and MRR was adopted by the challenge organizers to get the ranks of all participants.

   \subsection{Results}

   The results of our model on Ubuntu dataset and Advising dataset are summarized in Table~\ref{tab2}. We tuned our single models on the development datasets and submitted the final results for subtask 1 of the track using ensemble models. The ensemble models were built by averaging the outputs of three single models with identical architectures and different random initializations.
   % which is more accurate to represent the probability distribution. The best model was selected on the development dataset for testing. So we not only list the results with a single model and an ensemble model, but also list the results on the development dataset and on the test dataset here for reference.

   It should be noticed that the test set originally released for the Advising dataset had some dependency with the training set which we denoted as \emph{Advising-Case 1} in  Table~\ref{tab2}. %We also list the results on  this test set in Table~\ref{tab2}. To verify whether they have a strong ability to be generalized to the future conversations,
   The \emph{Advising-Case 2} test set was further released to better evaluate model performance for unseen conversations and was used for system ranking. % On Advising dataset, \emph{Advising-Case 2} results have been considered for ranking which is more persuasive.

   %The model we mentioned above has achieved a promising results that ranking second on the Ubuntu dataset and ranking third on the Advising dataset in the response selection track.
    According to the evaluation results released by challenge organizers, our proposed method ranked second on the Ubuntu dataset and third on the Advising dataset in subtask 1 of Track 1 among all 20 participants.

\section{Analysis}

   \subsection{Dataset comparison}
   From the evaluation results on the two different datasets shown in Table~\ref{tab2}, we can see that there were significant recall and MRR differences between the two datasets although the same model architectures were shared. We have mentioned above that these two datasets were different in the sizes of development set, test set and vocabulary.
   Although the Ubuntu dataset had a much larger vocabulary, its development/test set performances were better than the Advising dataset.
   %It is clear that our model is good at dealing with datasets with large vocabulary size. It has been proven that neural networks can learn better distribution representations given a lot of data. It is obvious that dataset with large vocabulary size can provide more relationships between words and furthermore can help represent sentence embeddings, which is possibly the reason for performing better on dataset with large vocabulary size.
   Meanwhile, our model showed a good generalization ability on the Ubuntu dataset because the evaluation results on test set were better than that on development set, showing its less dependency on the training set. However, the response selection performance on the Advising dataset was much worse.
   One possible reason is that the Advising dataset had a much small development set for model selection. Another reason is that there were some symbols such as \emph{EECS 351} and  \emph{Classes 280} which increased the difficulty of representation and modeling. % hard to represent and catch relationships with other courses.

   \subsection{Ablation tests}

   \begin{table}
     %\small
     \centering
     \begin{tabular}{lccc}
      \toprule
                  & Layer 1 & Layer 2 & Layer 3 \\
      \midrule
       Weights    & 0.5324  & 0.2067  & 0.2609  \\
      \bottomrule
      \end{tabular}
      \caption{Weights of each layer in AHRE}
      \label{tab4}
    \end{table}
   We further investigated the effects of different parts in our proposed model by removing them one by one. A single model built on the Ubutu dataset was adopted for this investigation and the development set performances are as shown in Table~\ref{tab3}.
   First, we can see that removing the modification layer degrades the recalls and MRR. This confirmed the positive effect of emphasizing the last utterance in the context for response selection. Second, we replaced the proposed AHRE with a simple single-layer BiLSTM at the encoding layer of our model and we can also see the performance degradation. Meanwhile, we also reported the learned weights of each layer in AHRE as shown in Table~\ref{tab4}.
   %This can be attributed to that %multi-layer RNN can encode the sequence at high level and
   %the attention mechanism over all RNN layers in AHRE incorporated more useful information such as morphology and semantics information than that from only the top layer.
   Furthermore, we replaced the multi-dimensional pooling and last-state pooling at the matching layer with max pooling and average pooling employed in original ESIM.
   The results shown that our proposed pooling strategy was more appropriate for the response selection task. %which can be interpreted as pooling at different granularities is more powerful to capture information.
   Finally, the word embedding were updated instead of being fixed during the training process, which also led to a performance degradation.
   Actually, the model described by the last row in Table~\ref{tab3} was the original ESIM.
   Comparing the first row and the last row in this table, we can see that significant performance improvement has been achieved by applying all our proposed techniques.
   %It can be seen that our model has improved the ESIM model significantly.

\section{Conclusion}
 In this paper, we have introduced our end-to-end model proposed for the response selection task in DSTC7. %showed the effectiveness of our proposed model in modelling response selection task by using the DSTC7 dataset.
 %Especially, this architecture performs well in enhancing the representation of words and sentences, and furthermore modifying matching degree with every candidate among the set. The word representation layer alleviates OOV problems very well. AHRE shows effectiveness on encoding the sentence at multi-level hierarchically and aggregate them to generate a more descriptive representations. The pooling method of combination of multi-dimentional pooling and last-state pooling forms the enhanced matching feature. The modification layer modifies the matching degree with every candidate among the set. The evaluation results show that our model has achieved a promising results of ranking second on the Ubuntu dataset and ranking third on the Advising dataset in the Track 1 of DSTC7. Due to the time limit, we focused on the subtask 1 becasue it is baseline and core of other subtasks. Several other subtasks are also very meaningful and helpful to the research community of dialogue.
 This model improves the original ESIM model from several aspects, including concatenated and fixed word representations, AHRE for sentence encoding, multi-dimentional  and last-state pooling for context-response matching, and score calculation with emphasis on the last utterance in the context.
 In the released evaluation results of DSTC7, our proposed method ranked second on the Ubuntu dataset and third on the Advising dataset in subtask 1 of Track 1 among all 20 participants. Ablation tests also confirm the effectiveness of our proposed methods.
 Our future work includes to explore the methods for other subtasks and to design a more domain-general framework that can alleviate domain-dependency of models.

%\section{Acknowledgments}

\bibliographystyle{aaai}
\bibliography{mybib}

\begin{thebibliography}{}

\bibitem[\protect\citeauthoryear{Abadi \bgroup et al\mbox.\egroup
  }{2016}]{abadi2016tensorflow}
Abadi, M.; Barham, P.; Chen, J.; Chen, Z.; Davis, A.; Dean, J.; Devin, M.;
  Ghemawat, S.; Irving, G.; Isard, M.; et~al.
\newblock 2016.
\newblock Tensorflow: a system for large-scale machine learning.
\newblock In {\em OSDI}, volume~16,  265--283.

\bibitem[\protect\citeauthoryear{Bahdanau, Cho, and
  Bengio}{2014}]{bahdanau2014neural}
Bahdanau, D.; Cho, K.; and Bengio, Y.
\newblock 2014.
\newblock Neural machine translation by jointly learning to align and
  translate.
\newblock {\em arXiv preprint arXiv:1409.0473}.

\bibitem[\protect\citeauthoryear{Chen \bgroup et al\mbox.\egroup
  }{2017}]{chen2017enhanced}
Chen, Q.; Zhu, X.; Ling, Z.-H.; Wei, S.; Jiang, H.; and Inkpen, D.
\newblock 2017.
\newblock Enhanced lstm for natural language inference.
\newblock In {\em Proceedings of the 55th Annual Meeting of the Association for
  Computational Linguistics (Volume 1: Long Papers)}, volume~1,  1657--1668.

\bibitem[\protect\citeauthoryear{Dong and Huang}{2018}]{dong2018enhance}
Dong, J., and Huang, J.
\newblock 2018.
\newblock Enhance word representation for out-of-vocabulary on ubuntu dialogue
  corpus.
\newblock {\em arXiv preprint arXiv:1802.02614}.

\bibitem[\protect\citeauthoryear{Hashimoto \bgroup et al\mbox.\egroup
  }{2017}]{hashimoto2017joint}
Hashimoto, K.; Tsuruoka, Y.; Socher, R.; et~al.
\newblock 2017.
\newblock A joint many-task model: Growing a neural network for multiple nlp
  tasks.
\newblock In {\em Proceedings of the 2017 Conference on Empirical Methods in
  Natural Language Processing},  1923--1933.

\bibitem[\protect\citeauthoryear{Hochreiter and
  Schmidhuber}{1997}]{hochreiter1997long}
Hochreiter, S., and Schmidhuber, J.
\newblock 1997.
\newblock Long short-term memory.
\newblock {\em Neural computation} 9(8):1735--1780.

\bibitem[\protect\citeauthoryear{Kingma and Ba}{2014}]{kingma2014adam}
Kingma, D.~P., and Ba, J.
\newblock 2014.
\newblock Adam: A method for stochastic optimization.
\newblock {\em arXiv preprint arXiv:1412.6980}.

\bibitem[\protect\citeauthoryear{Kummerfeld \bgroup et al\mbox.\egroup
  }{2018}]{kummerfeld2018analyzing}
Kummerfeld, J.~K.; Gouravajhala, S.~R.; Peper, J.; Athreya, V.; Gunasekara, C.;
  Ganhotra, J.; Patel, S.~S.; Polymenakos, L.; and Lasecki, W.~S.
\newblock 2018.
\newblock Analyzing assumptions in conversation disentanglement research
  through the lens of a new dataset and model.
\newblock {\em arXiv preprint arXiv:1810.11118}.

\bibitem[\protect\citeauthoryear{Mikolov \bgroup et al\mbox.\egroup
  }{2010}]{mikolov2010recurrent}
Mikolov, T.; Karafi{\'a}t, M.; Burget, L.; {\v{C}}ernock{\`y}, J.; and
  Khudanpur, S.
\newblock 2010.
\newblock Recurrent neural network based language model.
\newblock In {\em Eleventh Annual Conference of the International Speech
  Communication Association}.

\bibitem[\protect\citeauthoryear{Mikolov \bgroup et al\mbox.\egroup
  }{2013}]{mikolov2013distributed}
Mikolov, T.; Sutskever, I.; Chen, K.; Corrado, G.~S.; and Dean, J.
\newblock 2013.
\newblock Distributed representations of words and phrases and their
  compositionality.
\newblock In {\em Advances in neural information processing systems},
  3111--3119.

\bibitem[\protect\citeauthoryear{Nair and Hinton}{2010}]{nair2010rectified}
Nair, V., and Hinton, G.~E.
\newblock 2010.
\newblock Rectified linear units improve restricted boltzmann machines.
\newblock In {\em Proceedings of the 27th international conference on machine
  learning (ICML-10)},  807--814.

\bibitem[\protect\citeauthoryear{Pennington, Socher, and
  Manning}{2014}]{pennington2014glove}
Pennington, J.; Socher, R.; and Manning, C.
\newblock 2014.
\newblock Glove: Global vectors for word representation.
\newblock In {\em Proceedings of the 2014 conference on empirical methods in
  natural language processing (EMNLP)},  1532--1543.

\bibitem[\protect\citeauthoryear{Peters \bgroup et al\mbox.\egroup
  }{2018}]{peters2018deep}
Peters, M.; Neumann, M.; Iyyer, M.; Gardner, M.; Clark, C.; Lee, K.; and
  Zettlemoyer, L.
\newblock 2018.
\newblock Deep contextualized word representations.
\newblock In {\em Proceedings of the 2018 Conference of the North American
  Chapter of the Association for Computational Linguistics: Human Language
  Technologies, Volume 1 (Long Papers)}, volume~1,  2227--2237.

\bibitem[\protect\citeauthoryear{Serban \bgroup et al\mbox.\egroup
  }{2016}]{serban2016building}
Serban, I.~V.; Sordoni, A.; Bengio, Y.; Courville, A.~C.; and Pineau, J.
\newblock 2016.
\newblock Building end-to-end dialogue systems using generative hierarchical
  neural network models.

\bibitem[\protect\citeauthoryear{Shang, Lu, and Li}{2015}]{shang2015neural}
Shang, L.; Lu, Z.; and Li, H.
\newblock 2015.
\newblock Neural responding machine for short-text conversation.
\newblock In {\em Proceedings of the 53rd Annual Meeting of the Association for
  Computational Linguistics and the 7th International Joint Conference on
  Natural Language Processing (Volume 1: Long Papers)}, volume~1,  1577--1586.

\bibitem[\protect\citeauthoryear{Shen \bgroup et al\mbox.\egroup
  }{2017}]{shen2017disan}
Shen, T.; Zhou, T.; Long, G.; Jiang, J.; Pan, S.; and Zhang, C.
\newblock 2017.
\newblock Disan: Directional self-attention network for rnn/cnn-free language
  understanding.
\newblock {\em arXiv preprint arXiv:1709.04696}.

\bibitem[\protect\citeauthoryear{Wang, Hamza, and
  Florian}{2017}]{wang2017bilateral}
Wang, Z.; Hamza, W.; and Florian, R.
\newblock 2017.
\newblock Bilateral multi-perspective matching for natural language sentences.
\newblock In {\em Proceedings of the 26th International Joint Conference on
  Artificial Intelligence},  4144--4150.
\newblock AAAI Press.

\bibitem[\protect\citeauthoryear{Wu \bgroup et al\mbox.\egroup
  }{2017}]{wu2017sequential}
Wu, Y.; Wu, W.; Xing, C.; Zhou, M.; and Li, Z.
\newblock 2017.
\newblock Sequential matching network: A new architecture for multi-turn
  response selection in retrieval-based chatbots.
\newblock In {\em Proceedings of the 55th Annual Meeting of the Association for
  Computational Linguistics (Volume 1: Long Papers)}, volume~1,  496--505.

\bibitem[\protect\citeauthoryear{Zhou \bgroup et al\mbox.\egroup
  }{2016}]{zhou2016multi}
Zhou, X.; Dong, D.; Wu, H.; Zhao, S.; Yu, D.; Tian, H.; Liu, X.; and Yan, R.
\newblock 2016.
\newblock Multi-view response selection for human-computer conversation.
\newblock In {\em Proceedings of the 2016 Conference on Empirical Methods in
  Natural Language Processing},  372--381.

\end{thebibliography}

\end{document}